%% file: main.tex
\documentclass[lettersize,journal]{IEEEtran}
\usepackage{amsmath,amsfonts}
\usepackage{algorithmic}
\usepackage{algorithm}
\usepackage{array}
\usepackage[caption=false,font=normalsize,labelfont=sf,textfont=sf]{subfig}
\usepackage{textcomp}
\usepackage{stfloats}
\usepackage{url}
\usepackage{verbatim}
\usepackage{graphicx}
\usepackage{cite}
\usepackage{multirow}
\usepackage{color}
\usepackage{tikz}
\usepackage{hyperref}
\usetikzlibrary{arrows.meta,positioning,fit,shapes.multipart,calc}
\begin{document}

\title{Task-Routed Mixture-of-Experts with Cognitive Appraisal for Implicit Sentiment Analysis}

%%%%%%%%%%%%%%%%%  Below is the ORIGINAL content %%%%%%%%%%%%%%%

\author{Yaping Chai, Haoran Xie, Joe S. Qin
\thanks{This work was supported by the Research Impact Fund by the Research Grants Council of Hong Kong (Project No. 130272); a grant from the Research Grants Council of the Hong Kong Special Administrative Region, China (R1015-23); the Faculty Research Grants (SDS24A8, SDS25A15, and SDS24A19), Interdisciplinary \& Strategic Research Grant (ISRG252606), and the Direct Grants (DR25E8 and DR26F2) of Lingnan University, Hong Kong.
\emph{(Corresponding author: Haoran Xie.)}}

\thanks{Yaping Chai, Haoran Xie, and Joe S. Qin are with the Division of Artificial Intelligence, Lingnan University, Hong Kong (e-mail: yapingchai@ln.hk; hrxie@ln.edu.hk; joeqin@ln.edu.hk).}
}

%%%%%%%%%%%%%%% The content below needs to be deleted or modified %%%%%%%%%


%%%%%%%%%%%%%%%%%%%%%%%%%%%%%%%%%%%%%%%%%%%%%%%%%%%%%%%%%%%%%%%%%%%%%%%%%%%

\maketitle

\begin{abstract}
Implicit sentiment analysis is challenging because sentiment toward an aspect is often inferred from events rather than expressed through explicit opinion words. Existing models typically learn from the final polarity label, which provides limited guidance for reasoning about sentiment from the context. Motivated by cognitive appraisal theory, we propose an appraisal-aware multi-task learning (MTL) framework for implicit sentiment analysis that provides polarity prediction with two complementary auxiliary tasks: implicit sentiment detection and cognitive rationale generation. However, training several objectives with different targets and sharing a single backbone across tasks in MTL limits flexibility and can lead to task interference. To reduce interference among these related but distinct objectives, we adopt task-level mixture-of-experts models in which all tasks share a common set of experts, and task identity controls the sparse combination of these experts. Our method builds on an encoder-decoder architecture and replaces a subset of encoder and decoder blocks with these sparse mixtures. We use a task-conditioned router to select sparse expert mixtures for each task, and a task-separated routing objective to encourage different tasks to learn distinct expert-selection patterns. Experimental results show that our model outperforms recently proposed approaches, with strong gains on the implicit sentiment subset. Our code is available at \href{https://github.com/yaping166/TRMoE-ISA}{https://github.com/yaping166/TRMoE-ISA}.
 
\end{abstract}

\begin{IEEEkeywords}
Implicit sentiment analysis, mixture-of-experts, multi-task learning, large language models
\end{IEEEkeywords}

\input{1_introduction}

\input{2_related_work}

\input{3_method}

\input{4_exp}

\input{5_conclusion}

%\section*{Acknowledgment}
%This work was supported by the Research Impact Fund by the Research Grants Council of Hong Kong (Project No. 130272); a grant from the Research Grants Council of the Hong Kong Special Administrative Region, China (R1015-23); the Faculty Research Grants (SDS24A8, SDS25A15, and SDS24A19), Interdisciplinary \& Strategic Research Grant (ISRG252606), and the Direct Grants (DR25E8 and DR26F2) of Lingnan University, Hong Kong.

%%%%%%%%%%%% REFERENCES %%%%%%%%%%%%

% \input{bbl}   

\bibliographystyle{IEEEtran}
% \bibliography{IEEEabrv, main}
\bibliography{main}
\vspace{-20pt} 


\begin{IEEEbiography}[{\includegraphics[width=1in, height=1.25in,clip,keepaspectratio]{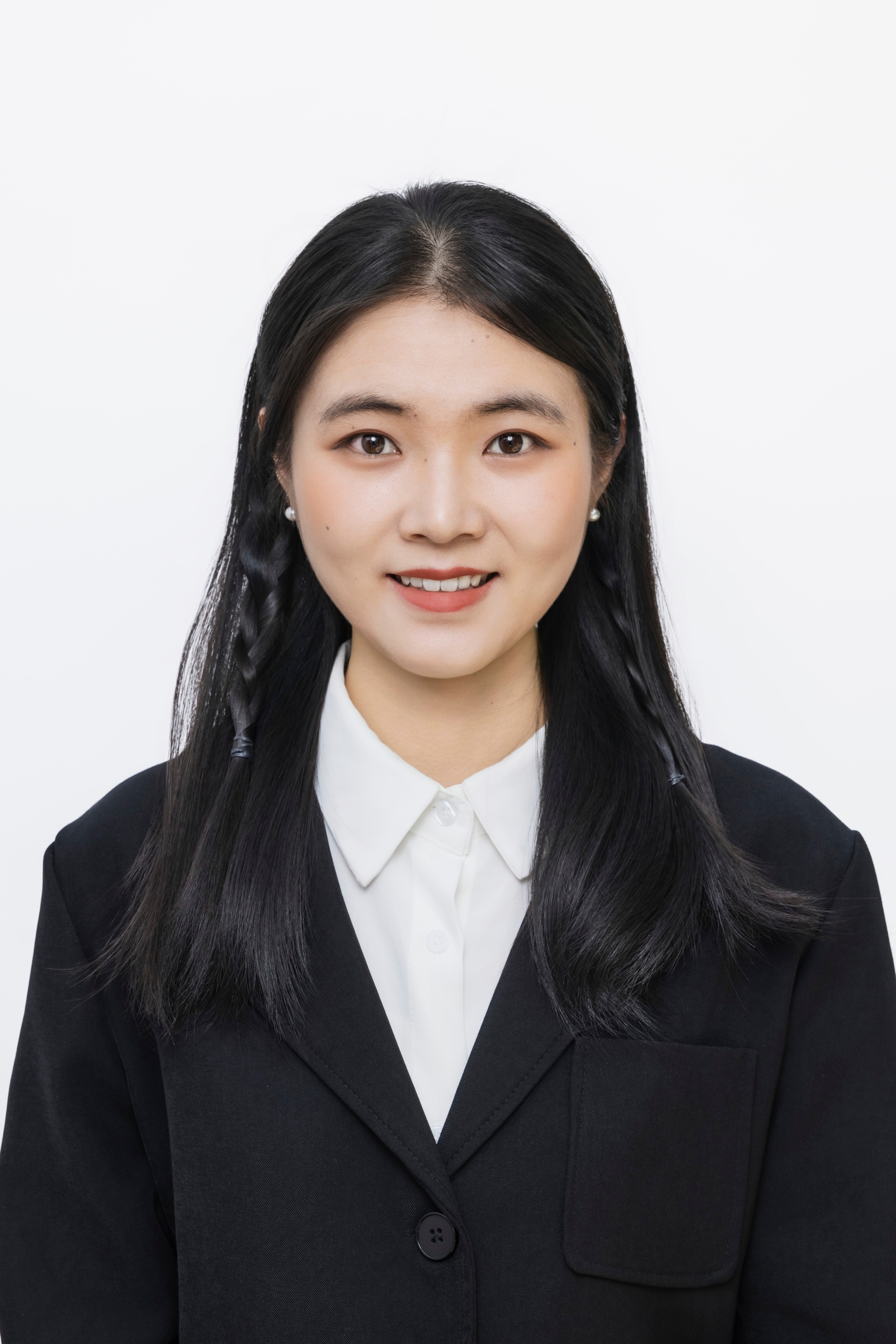}}]{Yaping Chai} is a Ph.D. candidate in the School of Data Science at Lingnan University, Hong Kong, supervised by Prof. Joe S. Qin and Prof. Haoran Xie. Her research interests center on large language models, natural language processing, and aspect-based sentiment analysis. She investigates advanced techniques for fine-tuning and assessing language models for sentiment analysis and other related NLP tasks.
\end{IEEEbiography}
\vspace{-25pt}

\begin{IEEEbiography}[{\includegraphics[width=1in,height=1.25in,clip,keepaspectratio]{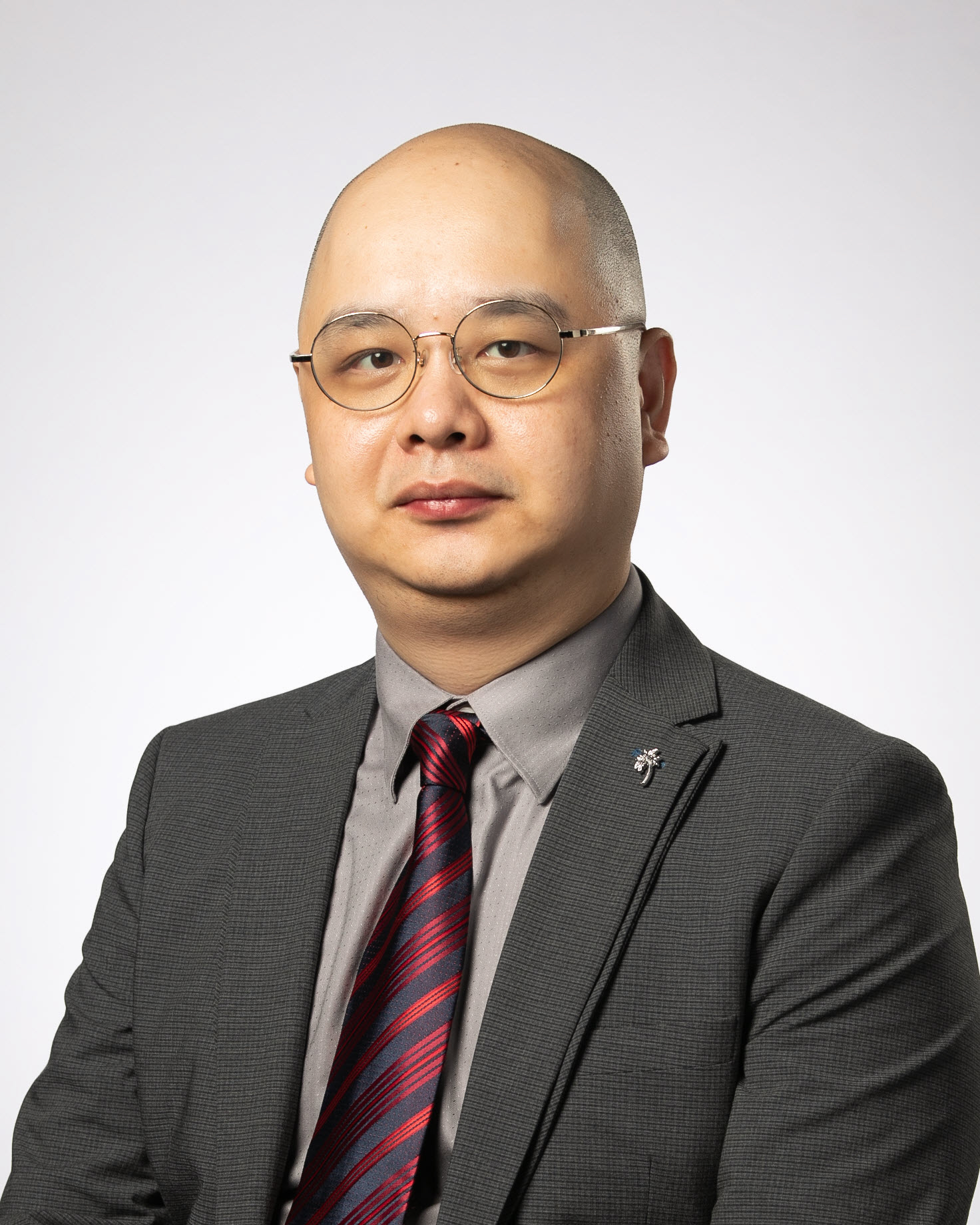}}]{Haoran Xie} (Senior Member, IEEE)
received a Ph.D. degree in Computer Science from City University of Hong Kong and an Ed.D degree in Digital Learning from the University of Bristol. He is currently a Professor and the Person-in-Charge at the Division of Artificial Intelligence, Director of LEO Dr David P. Chan Institute of Data Science, and Associate Dean of the School of Data Science, Lingnan University, Hong Kong. His research interests include natural language processing, large language models, language learning, and AI in education. He has published 500 research publications, including 300 journal articles. He is the Editor-in-Chief of Natural Language Processing Journal, Computers \& Education: Artificial Intelligence, and Computers \& Education: X Reality. He has been selected as the World's Top 2\% Scientists by Stanford University.
\end{IEEEbiography}
\vspace{-25pt}

% \vspace{-1em}
\begin{IEEEbiography}[{\includegraphics[width=1in,height=1.25in,clip,keepaspectratio]
{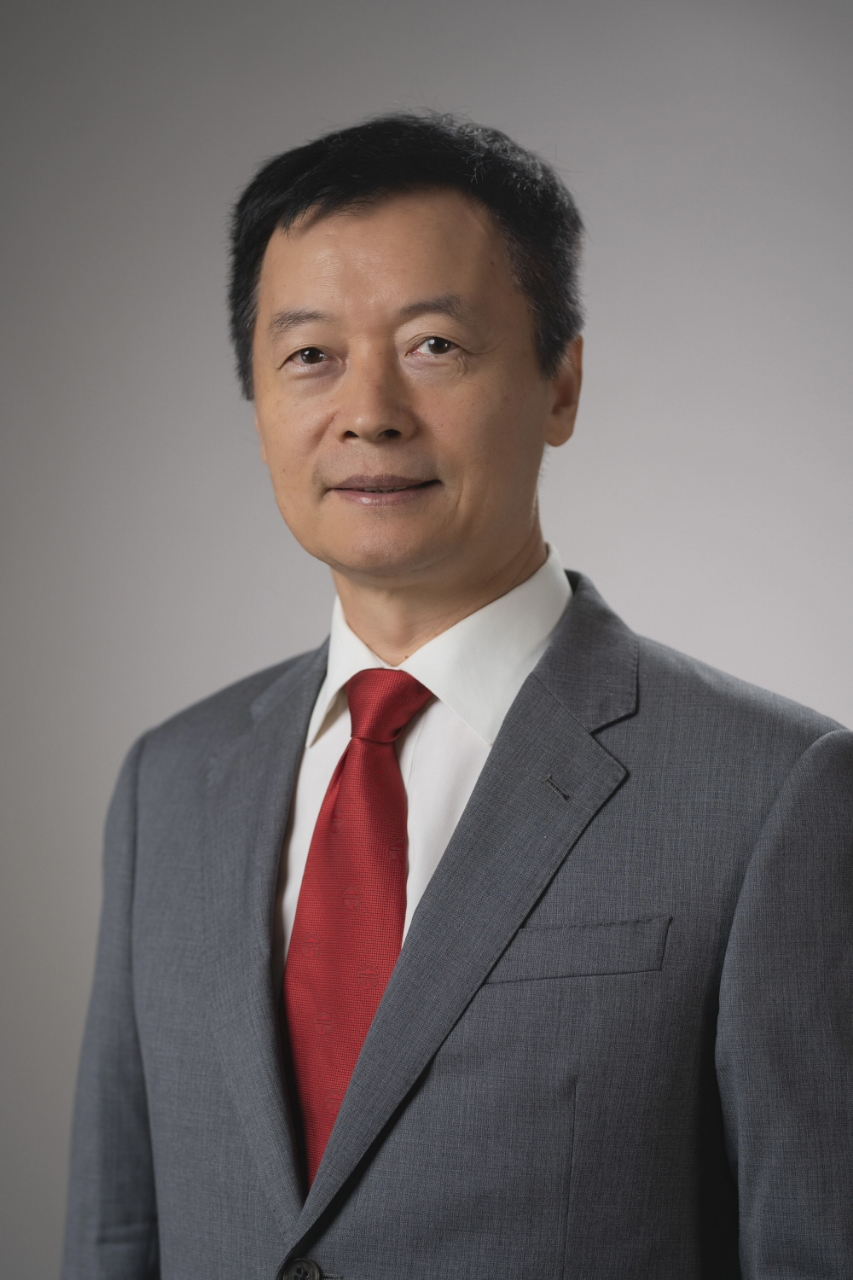}}]
{S. Joe Qin} (Fellow, IEEE) received the B.S. and M.S. degrees in automatic control from Tsinghua University, Beijing, China, in 1984 and 1987, respectively, and the Ph.D. degree in chemical engineering from the University of Maryland, College Park, MD, USA, in 1992. He is currently the Wai Kee Kau Chair Professor and President of Lingnan University, Hong Kong. His research interests include data science and analytics, machine learning, process monitoring, model predictive control, system identification, smart manufacturing, smart cities, and predictive maintenance. Prof. Qin is a Fellow of the U.S. National Academy of Inventors, IFAC, and AIChE. He was the recipient of the 2022 CAST Computing Award by AIChE, 2022 IEEE CSS Transition to Practice Award, U.S. NSF CAREER Award, and NSF-China Outstanding Young Investigator Award. His h-indices for Web of Science, SCOPUS, and Google Scholar are 66, 73, and 89, respectively.
\end{IEEEbiography}
\vfill

\end{document}

%% file: 1_introduction.tex
\section{Introduction}
\label{sec:introduction}

Aspect-based sentiment analysis (ABSA) aims to identify the sentiment
polarity expressed toward a specific aspect in a review and has become a
central research direction for understanding fine-grained opinions 
\cite{DL-absa}. However, much of the existing work assumes that sentiment evidence is provided in the text. Words such as ``delicious", ``rude", or ``overpriced" often indicate the speaker's evaluative intention. In contrast, the sentence ``We waited forty minutes before anyone came to the table" does not contain an obvious negative adjective. However, it still expresses dissatisfaction with the service. This scenario, known as implicit sentiment analysis (ISA), is challenging because polarity cannot be inferred solely on explicit sentiment words \cite{SCAPT}.

Existing approaches have made progress by improving contextual representations \cite{RGAT} or aligning implicit expressions with explicit sentiment words \cite{ABSA-ESA}. Nevertheless, most learning objectives are still centered on a single polarity label. This guidance tells the model what sentiment is expressed, but does not provide information about the reasoning behind the sentiment judgment. In implicit
cases, a reasoning chain is important for interpretation \cite{thor,cot_implicit}. For example, a consumer who receives the main course much later than
everyone else at the table may appraise the event as blocking the goal of
timely service (goal inconduciveness) and violating expected service
norms (norm incompatibility) \cite{reasoning_chain}. These appraisals explain how speakers evaluate events along core psychological dimensions, e.g., goals and expectations, before
a negative attitude toward service emerges. These intermediate cognitive steps explain why attitudes toward service are
likely to be negative, yet single-task polarity learning only provides the final polarity outcome (negative), rather than the intermediate steps that led to it, which limits the learning signal for implicit sentiment reasoning.

According to cognitive appraisal theory, affective responses arise from the cognitive evaluation of psychological factors, including goals, expectations, agency, and consequences \cite{lazarus1991emotion,scherer2001appraisal}. This perspective suggests that aspect-level sentiment is not only a polarity decision but also the outcome of affective processes. Motivated by this view, we formulate ISA as an appraisal-aware multi-task learning problem. In addition to polarity classification, we introduce cognitive appraisal reasoning, in which the model generates a brief explanation of why the speaker holds an attitude toward the target aspect. The rationale provides intermediate affective information that links the event in the review to the final polarity. We also introduce implicit sentiment detection, where the model predicts whether the sentiment evidence is explicit or implicit. This task guides the model to look beyond lexical sentiment words and infer the affective implication of events.

These auxiliary tasks are complementary to polarity classification. However, they require different targets: polarity classification focuses on the final sentiment label, implicit sentiment detection focuses on an explicit-implicit evidence judgment, and rationale generation requires explanatory sentence generation. In multi-task learning, it is common to use a single backbone across tasks with different formats and objectives, which can lead to task interference and negative transfer when parameters are updated jointly \cite{standley2020task,MTL_mitigating}. Sharing affective knowledge across related objectives while limiting interference among them remains challenging.

Task-level mixture-of-experts (MoE) provides an efficient way to address this challenge \cite{moe_survey,moe_survey2}. Prior task-level MoE studies \cite{modsquad,crosstaskmoe,tacomoe} show that expert routing can help multi-task models support both cooperation and specialization by assigning different tasks to different expert mixtures while still allowing shared experts when tasks need overlapping knowledge. This motivates us to use a task-routed MoE architecture, where the router is conditioned on task identity, allowing different tasks to use task-specific expert mixtures rather than passing through the same network at every layer.

Inspired by the above motivations, we propose a unified framework for implicit sentiment analysis. It formulates each aspect-level instance into three natural-language tasks: polarity classification, implicit sentiment detection, and cognitive appraisal reasoning. Our method
builds on a sequence-to-sequence backbone \cite{flan-t5} with task-routed expert layers. Each task has a learnable representation, and each routed layer uses this representation to select a sparse mixture of feed-forward experts. We further introduce a task-separated routing objective. This objective reduces the similarity between the gate distributions of different tasks, encouraging them to form separable routing patterns across expert layers. The model therefore preserves shared pre-trained language knowledge while creating task-conditioned pathways for appraisal-aware implicit sentiment inference.

Our contributions are as follows:
\begin{itemize}
    \item We adopt task-routed mixture-of-experts for appraisal-aware implicit sentiment analysis and use task identity to route samples, enabling related tasks to share knowledge while keeping their expert selection patterns different.
    \item We propose a task-separated routing objective that encourages
    different tasks to acquire separable expert-selection patterns, reducing task interference.
    \item Empirical evaluation of the benchmarks shows that our method achieves a strong performance against recent methods, including implicit sentiment
 subsets, demonstrating the effectiveness of the proposed approach.
\end{itemize}

%% file: 2_related_work.tex
\section{Related Work}
\label{sec:related-work}

\subsection{Implicit Sentiment Analysis}

Implicit sentiment analysis focuses on situations in which opinion words do not directly express sentiment toward an aspect. Prior work addresses the scarcity of implicit sentiment expressions by aligning representations or using external knowledge and synthetic data. For example, \cite{SCAPT} introduces supervised
contrastive pre-training for implicit sentiment, using contrastive learning,
review reconstruction, and masked aspect prediction to align representations of
explicit and implicit sentiment expressions. \cite{C3DA} constructs
multi-aspect samples with aspect and polarity augmentation channels, and uses
contrastive learning with an entropy-based filter to reduce noise from generated examples. Recent text data augmentation methods enhance model performance by leveraging the generative capabilities of large language models \cite{da}. For example, \cite{thor} uses
chain-of-thought prompting to induce the latent aspect, the underlying opinion, and the final polarity step by step. Other studies utilize the internal logic and syntactic dependencies that connect aspects to their implicit sentiments. For example, \cite{RGAT} proposes a relational graph attention network that uses an aspect-oriented dependency tree to reshape syntactic structure around
the target aspect, helping the model connect aspects with their relevant
opinion expressions. \cite{ISAIV} studies the influence of confounding
sentiment words and uses instrumental variables with stochastic perturbations to
estimate a cleaner causal relation between a sentence and its sentiment. These methods show that ISA benefits from richer context modeling and implicit-explicit alignment. However, most of them still optimize the model mainly toward the final polarity decision. In contrast, our work treats implicitness detection and cognitive rationale
generation as auxiliary tasks, so that auxiliary objectives complement the polarity classification with additional supervision dimensions beyond surface sentiment labels.

\subsection{Multi-task Learning for ABSA}

Multi-task learning enables knowledge sharing across related tasks, 
allowing the model to leverage complementary information from 
different tasks \cite{MTL}. In ABSA, \cite{BERT-ADA} improves
aspect-target sentiment classification through domain-specific language
model fine-tuning followed by task-specific supervised training, showing that
domain-aware auxiliary training can reduce the mismatch between general
pre-training and target-domain sentiment prediction.
\cite{ABSA-ESA} generates explicit sentiment
augmentations for implicit cases and integrates them as additional clues for
polarity prediction. \cite{laimulti} studies multi-task implicit sentiment analysis with large language models, constructing auxiliary sentiment-element tasks and using automatic weight learning to handle
data and task uncertainty. However, training several objectives with different targets and sharing a single backbone across tasks in MTL limits flexibility and can lead to task interference \cite{standley2020task}. Task-level mixture-of-experts mitigates this by maintaining a single set of experts and activating different experts for each task. For example, \cite{crosstaskmoe} uses task representations to route tasks through different expert combinations and analyze the learned cross-task skills. \cite{modsquad} uses MoE layers and a mutual-information objective to encourage sparse dependencies between tasks and experts, balancing cooperation and specialization in multi-task learning. Beyond task-conditioned sparse experts, we propose a task-separated routing objective that allows related tasks to share knowledge while enabling different tasks to select distinct experts, reducing task interference.

%% file: 3_method.tex
\section{Method}
\label{sec:method}

\begin{figure*}[htbp]
    \centering
    \includegraphics[width=\textwidth]{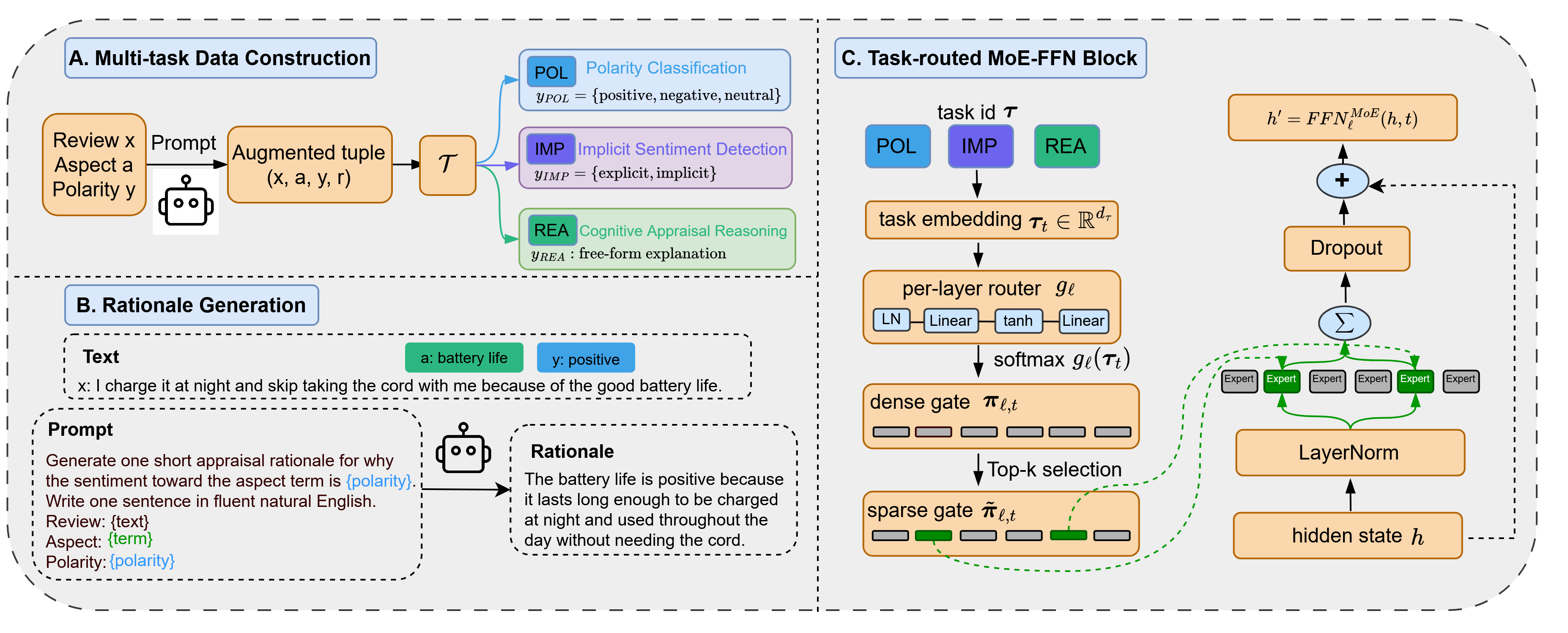}
    \caption{Overview of our framework. A. Multi-task Data Construction: each aspect-level instance is formulated into three text-to-text tasks. B. Rationale Generation: Prompt template for the cognitive appraisal rationale generation task, along with an example of the model output. C. Task-routed MoE-FFN Block: layer normalization is applied before the experts; only the top-k expert weights are kept; the outputs of those experts are then combined using the routing weights.}
    \label{fig:overview}
\end{figure*}
Our framework has three components. First, we convert each aspect-level instance into a unified text-to-text multi-task
problem containing polarity classification, implicit sentiment detection, and
cognitive appraisal reasoning. Second, we replace the feed-forward modules in selected encoder and decoder layers with a set of task-conditioned experts, allowing the model to share general linguistic knowledge while assigning different expert mixtures based on the task identifier. Third, we use a task-separated routing objective to encourage the learned routing patterns
to be separable across tasks. Fig. \ref{fig:overview} shows the overview of our method.

\subsection{Problem Formulation}
\label{sec:method-formulation}

Each annotated instance is a tuple
$(x, a, y, e, r)$, where $x\in\mathcal{X}$ is a review sentence; $a$ is an
aspect term in $x$; $y\in\mathcal{Y}=\{\text{positive},\text{negative},\text{neutral}\}$ is the
gold polarity toward $a$. The dataset provides an implicitness indicator
$e\in\{0,1\}$, with $e{=}1$ for implicit sentiment and $e{=}0$ for explicit
sentiment, together with a short rationale $r$ that explains why $y$ is
assigned to $(x,a)$. Instead of only predicting the polarity $y$, we formulate each aspect-level instance as
three text-to-text tasks:
\[
\mathcal{T}=\{\textsc{pol},\textsc{imp},\textsc{rea}\}
\]
covering polarity classification (\textsc{pol}), implicitness detection
(\textsc{imp}), and cognitive appraisal reasoning (\textsc{rea}). Training on all three tasks yields supervision at different targets: the sentiment
decision, whether the sentiment is expressed via explicit opinion cues or inferred from context, and a textual description of the underlying appraisal.

\subsection{Cognitive Appraisal Rationale Generation}
\label{sec:method-appraisal}

In order to improve the reasoning capability of ISA language models, we use a large language model (LLM) to generate a cognitive appraisal rationale that explicitly links the sentence, the target term, and the sentiment polarity. The LLM generates an appraisal-based rationale \(r\) for cognitive evaluations, indicating the relationship between \((x, a)\) and \(y\), creating an augmented element set \((x, a, y, r)\) that can be used to train models to perform ISA with appraisal reasoning. Fig. \ref{fig:overview}(B) shows the prompt template and an example output. These rationales enable the model to integrate appraisal-based information for implicit sentiment analysis.

\subsection{Task-Routed Mixture-of-Experts}
\label{sec:method-trmoe}

\paragraph{Expert Set}
We adopt a pretrained encoder-decoder language model \cite{flan-t5} containing $L$ transformer blocks. Each block contains an attention sublayer and a feed-forward sublayer $\mathrm{FFN}$. Following the previous sparse MoE design \cite{fedus2022switch}, we only modify layers
in $\mathcal{M}\subset\{1,\dots,L\}$. For each
$\ell\in\mathcal{M}$, the dense $\mathrm{FFN}_\ell$ in the corresponding
encoder and decoder blocks is replaced with $N$ parallel experts
$E_\ell^{(1)},\dots,E_\ell^{(N)}$, where $N$ represents the number of experts in that layer. Each $E_\ell^{(i)}$ keeps the same architecture as the
original $\mathrm{FFN}_\ell$.

\paragraph{Task-conditioned Router}

In contrast to the token-level routers of large language
MoEs \cite{fedus2022switch,shazeer2017moe,Uni-moe}, our router operates at the task
level, where the task identity $t$ determines the expert mixture. The three
tasks, $\textsc{pol}$, $\textsc{imp}$, and $\textsc{rea}$, are encoded as task IDs, and each task ID is associated with a task embedding
$\boldsymbol{\tau}_t\in\mathbb{R}^{d_\tau}$. For each routed layer
$\ell\in\mathcal{M}$, we use a separate gating network $g_\ell$ to map this
task embedding to $N$ expert logits:
\begin{equation}
g_\ell(\boldsymbol{\tau}_t)
= \mathrm{MLP}_\ell\!\bigl(\mathrm{LN}(\boldsymbol{\tau}_t)\bigr)
\in\mathbb{R}^{N}
\label{eq:router-logits}
\end{equation}
where $\mathrm{MLP}_\ell$ is a two-layer feed-forward network with a $\tanh$
activation. The expert gate is then
obtained by normalizing the logits:
\begin{equation}
\boldsymbol{\pi}_{\ell,t}
= \mathrm{softmax}\!\bigl(g_\ell(\boldsymbol{\tau}_t)\bigr)
\label{eq:gate}
\end{equation}
Thus $\boldsymbol{\pi}_{\ell,t}$ is the routing weight assigned to the $N$
experts for task $t$ at layer $\ell$.

\paragraph{Sparse Top-$k$ Selection}

We use the top-$k$ gating in sparse MoE layers. For task $t$ at
layer $\ell$, we only keep the $k$ experts with the largest probabilities in
$\boldsymbol{\pi}_{\ell,t}$, and $\tilde{\boldsymbol{\pi}}_{\ell,t}$ denote the resulting sparse gate after re-normalizing the retained weights. Let $\mathbf{h}$ represent the hidden state after the preceding sublayer, $\mathrm{LN}(\cdot)$ is layer normalisation, each $\{E_\ell^{(j)}\}_{j=1}^{N}$ is one expert feed-forward network in this layer, and $\mathrm{Drop}(\cdot)$ is dropout.
We compute $\mathrm{LN}(\mathbf{h})$, mix the expert outputs with
$\tilde{\boldsymbol{\pi}}_{\ell,t}$, add the mixture to $\mathbf{h}$, and apply
$\mathrm{Drop}(\cdot)$:
\begin{equation}
\mathrm{FFN}^{\mathrm{MoE}}_\ell(\mathbf{h},t)
\;=\;\mathbf{h}\;+\;\mathrm{Drop}\!\left(
\sum_{j=1}^{N}\tilde{\pi}_{\ell,t}^{(j)}\;
E_\ell^{(j)}\!\bigl(\mathrm{LN}(\mathbf{h})\bigr)
\right)
\label{eq:moe-output}
\end{equation}

\subsection{Learning Objectives}
\label{sec:method-objective}
The objectives below are defined on the encoder-decoder forward pass of $p_\theta$, where $\theta$ denotes all trainable parameters and $p_\theta$ is
the distribution over target tokens defined by the model with parameters $\theta$. At layers $\ell\in\mathcal{M}$, the dense feed-forward sublayer is
replaced by $\mathrm{FFN}^{\mathrm{MoE}}_\ell$ (Eq.~(\ref{eq:moe-output})).

\paragraph{Multi-task generation objective}
All three tasks are trained in a unified generative manner. For each task
$t \in \mathcal{T}=\{\textsc{pol},\textsc{imp},\textsc{rea}\}$, given the review-aspect pair under the corresponding task instruction, the model generates the task-specific target sequence
$y_t$. The multi-task generation objective is the standard sequence-to-sequence
negative log-likelihood:
\begin{equation}
\mathcal{L}_{\mathrm{gen}}(\theta)
\;=\;
-\sum_{t\in\mathcal{T}}
\mathbb{E}_{(x,a,y_t)\in\mathcal{D}_t}
\sum_{s=1}^{|y_t|}
\log p_\theta\!\left(y_{t,s} \mid y_{t,<s}, x, a, t\right)
\label{eq:lgen}
\end{equation}
where $y_{t,s}$ is the $s$-th token of the target sequence for task $t$,
$\mathcal{D}_t$ is the training set for task $t$, and $\theta$ denotes the
trainable parameters.

\paragraph{Task-separated routing objective}
To reduce task interference among related yet distinct objectives, we propose a routing objective that pushes the gate distributions of different tasks apart. 
Each routed layer $\ell\in\mathcal{M}$ produces
one dense gate vector $\boldsymbol{\pi}_{\ell,t}$ for each task
$t\in\mathcal{T}$. Let
$\mathcal{P}=\{(t,t'):t,t'\in\mathcal{T},t\neq t'\}$ denote the ordered task
pairs. We minimize the average cosine similarity between their gates:
\begin{equation}
\mathcal{L}_{\mathrm{sep}}(\theta)
\;=\;
\frac{1}{|\mathcal{M}|}\sum_{\ell\in\mathcal{M}}\,
\frac{1}{|\mathcal{P}|}
\sum_{(t,t')\in\mathcal{P}}
\cos\!\left(
\boldsymbol{\pi}_{\ell,t},\,
\boldsymbol{\pi}_{\ell,t'}
\right)
\label{eq:lsep}
\end{equation}
where
$\cos(\mathbf{u},\mathbf{v})=\mathbf{u}^{\top}\mathbf{v}/
(\lVert\mathbf{u}\rVert_2\lVert\mathbf{v}\rVert_2)$ is the cosine similarity. The dense gate $\boldsymbol{\pi}_{\ell,t}$ is the softmax-normalised output of the layer-specific router in Eq.~(\ref{eq:gate}), with logits $g_\ell(\boldsymbol{\tau}_t)$ given in Eq.~(\ref{eq:router-logits}).

\paragraph{Overall objective}
The final training objective combines generation loss and task-separated routing loss:
\begin{equation}
\mathcal{L}(\theta)
\;=\;
\mathcal{L}_{\mathrm{gen}}(\theta)
\;+\;
\lambda_{\mathrm{sep}}\,\mathcal{L}_{\mathrm{sep}}(\theta),
\label{eq:total}
\end{equation}
where $\lambda_{\mathrm{sep}}\ge 0$ controls the strength of the impact of routing separation. When $\lambda_{\mathrm{sep}}=0$, the router is trained only
through the generation objective.

%% file: 4_exp.tex
\section{Experiment}

\subsection{Datasets and Metrics}
We conduct experiments on the SemEval-2014 Restaurant and Laptop datasets \cite{SemEval}. Following
\cite{SCAPT}, we further divide each benchmark into explicit and implicit sentiment subsets according to whether the sentiment polarity toward opinion words directly expresses the target aspect. We reserve 10\% of the original
training set as the validation set and train the model on the remaining
training instances. We evaluate performance with respect to both overall and implicit sentiments.
Specifically, we report accuracy and macro-F1 score on the full test set, and accuracy on the implicit sentiment polarity subset. 

\subsection{Baselines}

We compare our method with three groups of baselines. 
\begin{itemize}
    \item \textbf{Conventional ABSA baselines:} these methods typically treat aspect-based sentiment analysis as a supervised classification
task. The models we use for comparison include BERT+SPC \cite{BERT}, BERT+ADA \cite{BERT-ADA}, RGAT \cite{RGAT}, BERT$_{Asp}$+CEPT \cite{SCAPT}, BERT$_{Asp}$+SCAPT \cite{SCAPT}, C3DA \cite{C3DA}, and ISAIV \cite{ISAIV}.
     \item \textbf{Inference-only baselines directly:} these methods use the zero-shot learning ability of the LLM, which prompts the LLM to classify each test instance
without task-specific training on the ABSA datasets. The baseline models include GPT-5.4-mini \cite{gpt-5},
DeepSeekV3.2 \cite{deepseek-v3.2} and Llama-3.3-70B-Instruct \cite{llama3}.
    \item \textbf{Instruction-based fine-tuning baselines:} these methods formulate the task as a generation problem. In the ABSA task, provide instruction prompts to the model and fine-tune it to generate the
target sentiment output. For comparison, we use ABSA-ESA \cite{ABSA-ESA}, Flan-T5 \cite{flan-t5}, InstructABSA \cite{Instructabsa}, THOR-prompt \cite{thor},
THOR \cite{thor}, and MT-ISA \cite{laimulti} as baselines.

\end{itemize}

\subsection{Models and Hyperparameters}

We use Flan-T5-large as the backbone model and use GPT-5.4-mini for cognitive appraisal
rationales generation. We replace selected encoder and decoder FFN
layers with task-routed MoE layers in layers 8, 10, 12, 14, 16, 18, 20, and
22. Each layer contains 5 experts and activates the top 2 experts for
each task. The routing separation weight $\lambda_{\mathrm{sep}}$ is set to 0.4 because the validation set achieves optimal performance at this value, as discussed in section \ref{lambda}. We train with a batch size of 4 and accumulate gradients in 4 steps, and the learning rate is $3\times
10^{-5}$. All reported results are the average of three runs with different random seeds.

\subsection{Main Results}

\input{table/ABSA}

\paragraph{Analysis of conventional ABSA baselines}
As shown in Table \ref{tab:main-results}, conventional ABSA baselines are not competitive in overall performance, especially in the implicit sentiment subset. These methods mainly train the model to predict the polarity label, relying on explicit opinion words when present. However, they do not guide the model to detect implicit evidence or explain why an event supports a sentiment. Instruction-based fine-tuning improves conventional ABSA baselines across both domains, with larger performance gains on the implicit subset, which is a particular challenge for ISA.

\paragraph{Analysis of inference-only baselines}
The inference-only baselines show that LLMs can leverage their pretrained knowledge to interpret sentiment reviews without task-specific training. However, zero-shot reasoning remains limited, and under the same prompting settings, stronger LLMs tend to score higher. For example, the GPT baseline ranks above the Llama baseline across all evaluation metrics and datasets. Despite the LLM having strong general reasoning ability, instruction-based fine-tuning, including our method, also improves over inference-only LLMs. This improvement suggests that aspect-specific supervision aligns the model with the benchmark and trainable inference skills. %Our auxiliary tasks add supervision for implicit and explicit detection and short rationale, which helps in situations where the review appears neutral on the surface but the aspect sentiment is implicit.

\paragraph{Analysis of instruction-based fine-tuning baselines}
In the instruction-based fine-tuning setup, smaller instruction-tuned models have benefited from the generative formulation. Still, they lack sufficient capacity to leverage the auxiliary supervision introduced in our framework. We observe that increasing the backbone to Flan-T5-large further improves this group's performance. With the Flan-T5-large backbone, our method achieves the best results, indicating that the proposed multi-task learning framework becomes more effective when the backbone has sufficient representation capacity. Compared with prior instruction-tuned methods, our method improves over this group because polarity prediction is trained together with implicitness detection and cognitive
appraisal reasoning, allowing the model to learn not only the lexical polarity cues but
also the latent evaluation process that supports the sentiment. The routing mechanism then allows these objectives to share general language knowledge while preserving task-specific expert pathways. 

\subsection{Ablation Study}

We perform ablation experiments to verify the impact of multi-task learning and the MoE routing architecture. As shown in Table \ref{tab:ablation}, Ours (w/o MTL) keeps the MoE architecture and trains only the polarity prediction task, and Ours (w/o MoE) removes the MoE layers and performs multi-task learning with the standard Flan-T5-large architecture. Removing multi-task learning (w/o MTL) degrades performance on both datasets. The larger decrease on ISA$_A$ indicates that implicit sentiment prediction benefits from auxiliary supervision of implicitness detection and cognitive appraisal rationale generation. Without these auxiliary tasks, the model mainly learns the direct mapping from the review and aspect to the polarity label, which weakens its ability to capture latent sentiment evidence. Removing the MoE architecture (w/o MoE) also leads to clear performance degradation, especially on Rest14, where ISA$_A$ decreases from 75.9\% to 71.16\%. This performance drop shows that simply applying multi-task learning with a shared Flan-T5-large backbone is insufficient to fully exploit different supervision signals. The task-routed MoE layers enable different tasks to share general linguistic knowledge while maintaining task-specific expert pathways, thereby improving both overall sentiment classification and implicit sentiment inference.

\input{table/ablation}

\subsection{Further Analysis}

\paragraph{Effect of $\lambda_{\mathrm{sep}}$}
\label{lambda}
\begin{figure*}[htbp]
    \centering
    \includegraphics[width=0.7\textwidth]{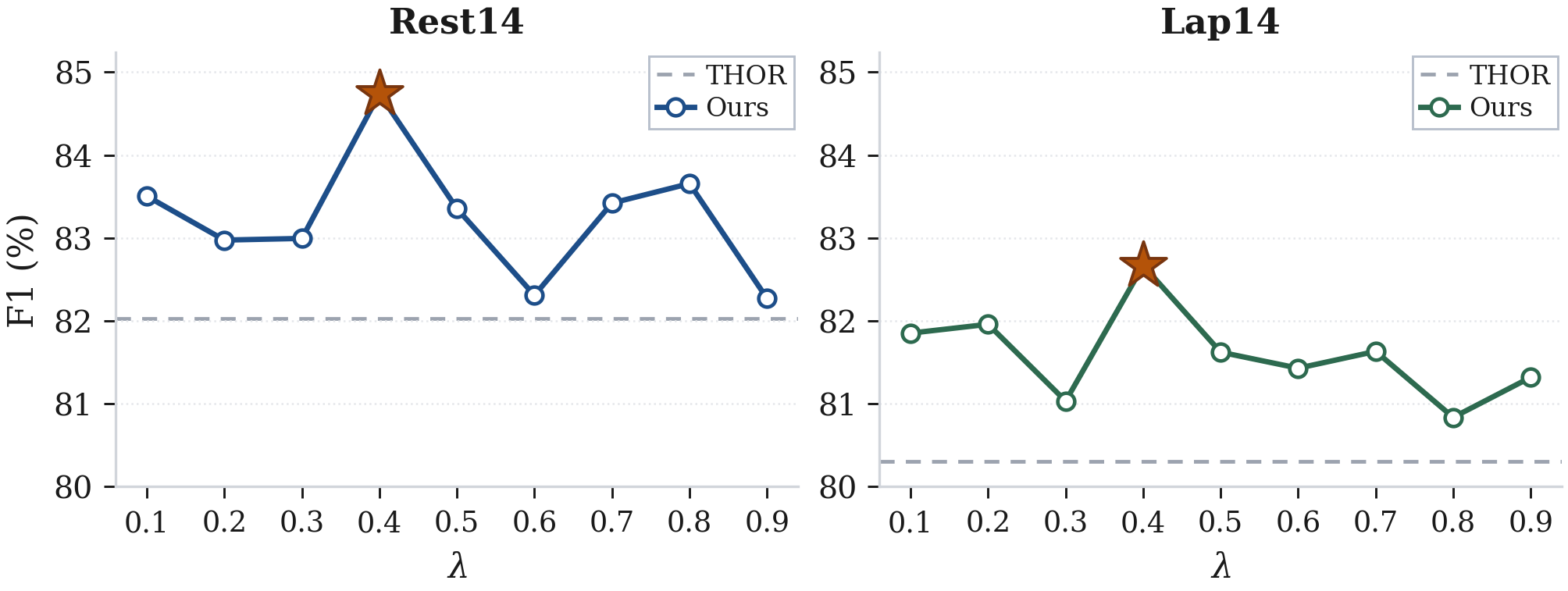}
\caption{Effect of the hyperparameter $\lambda_{\mathrm{sep}}$ on two benchmarks, where performance is measured by F1 score. Both datasets achieve their best performance at $\lambda_{\mathrm{sep}}=0.4$, and all $\lambda$ values remain above the THOR baseline. We mark the highest value with a $\star$.}
    \label{fig:lambda}
\end{figure*}

We study how the hyperparameter $\lambda_{\mathrm{sep}}$ affects experimental outcomes. As shown in Fig. \ref{fig:lambda}, our method outperforms THOR \cite{thor} in all $\lambda$ settings, demonstrating the effectiveness of our method. Both benchmarks achieve their best performance at an intermediate value of $\lambda_{\mathrm{sep}}=0.4$, indicating that a balanced routing separation weight improves model performance. If $\lambda_{\mathrm{sep}}$ is too small, tasks still overshare experts, and weaken task-specific signals; if it is too large, routing becomes overly constrained, weakening useful cross-task transfer.

\paragraph{Routing entropy across tasks}
\label{Routing_entropy}To study whether different tasks use MoE routing in different ways, we analyze routing entropy for implicitness detection (\textsc{imp}), polarity classification (\textsc{pol}), and cognitive appraisal rationale generation (\textsc{rea}), and compare these entropies under different routing separation weights $\lambda$. Routing entropy reflects how the model activates experts for each task: higher values indicate broader mixing across experts, while lower values indicate more concentrated, sparse expert usage.

\begin{figure*}[htbp]
    \centering
    \includegraphics[width=0.7\textwidth]{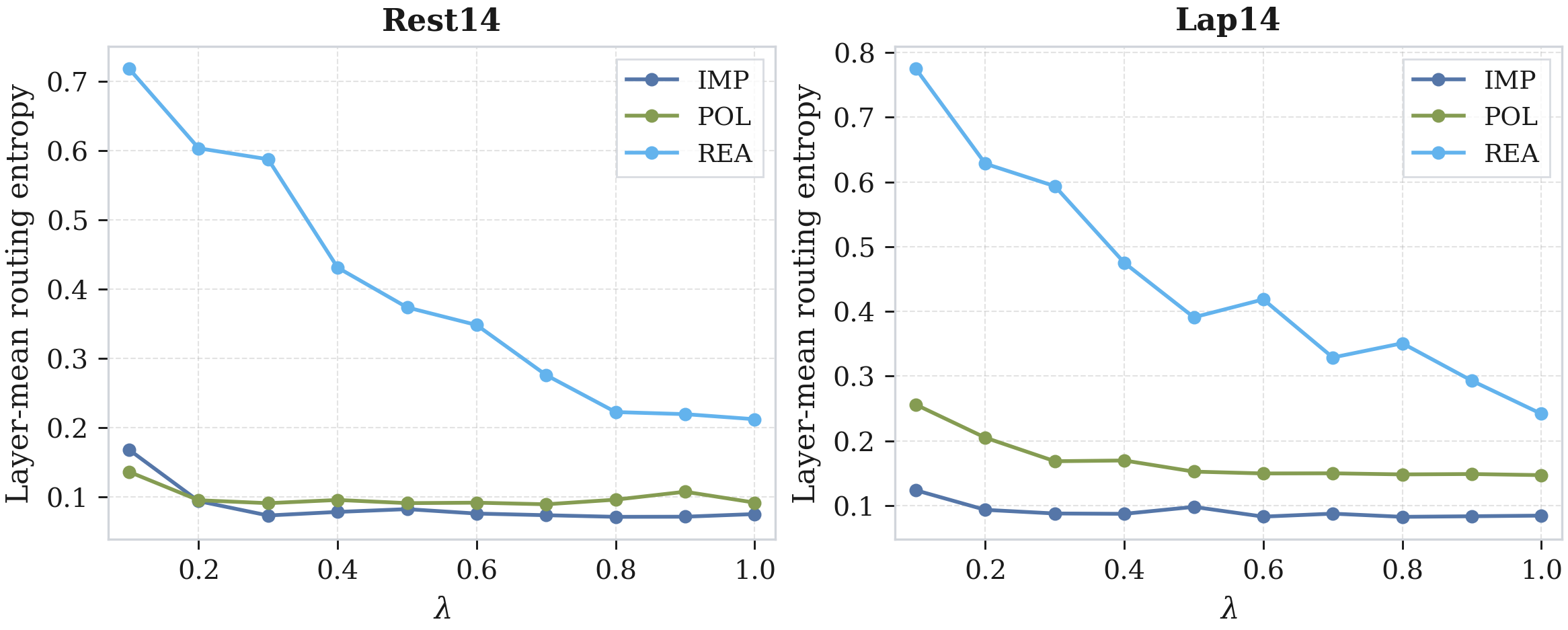}
    \caption{Routing entropy $H(\cdot)$ for the three tasks on Rest14 and Lap14 under different $\lambda$. Each point is the mean over routed MoE layers.}
    \label{fig:entropy}
\end{figure*}

As shown in Fig.~\ref{fig:entropy}, as $\lambda$ increases, the \textsc{rea} curve decreases, indicating that a larger routing separation weight leads to routing on fewer experts, thereby reducing task entropy. Additionally, across all $\lambda$ settings and in both domains, we observe a consistent phenomenon:
$H(\textsc{imp}) \approx H(\textsc{pol}) \ll H(\textsc{rea})$, where $H(\cdot)$ denotes routing entropy. The entropy pattern indicates that easier targets, such as classification and detection, can be achieved with a small subset of experts, whereas rationale generation demands richer expert mixing.

\begin{figure}[htbp]
    \centering
    \includegraphics[width=0.5\textwidth]{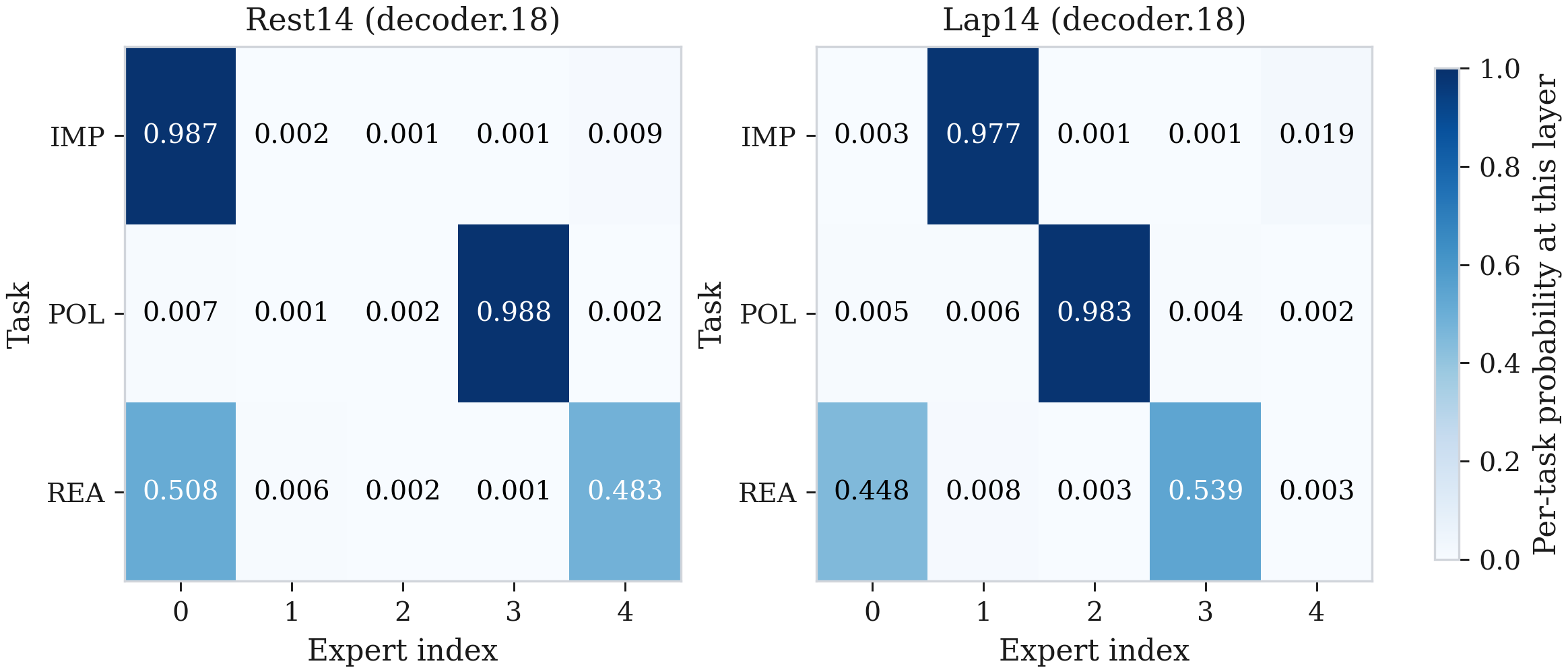}
    \caption{An example of the routing probabilities heatmap at the MoE decoder's 18th layer, where each row represents one of three tasks, each column represents an expert, and each cell represents the probability that the task assigns to that expert at this layer.}
    \label{fig:prob}
\end{figure}

\paragraph{Specialization of expert assignments}
We measure expert specialization by selecting the expert with the highest probability ($\text{Top-1}$) among the five experts in each routed layer, reflecting the dominance of the most active expert in each context. As shown in Fig. \ref{fig:prob}, for simpler detection and classification tasks, the $\text{Top-1}$ values across all MoE layers are notably high, indicating that a single expert dominates the activation. For the complex sequence generation task \textsc{rea}, the $\text{Top-1}$ is lower and can decrease further in certain layers, reflecting more mixed expert usage. This pattern aligns with the \textsc{rea} objective, which involves generating rationales that benefit from combining multiple experts rather than relying primarily on a single expert. The same phenomenon also occurs in the routing entropy discussed in section \ref{Routing_entropy}, where the entropy of \textsc{imp} and \textsc{pol} is lower, while the entropy of \textsc {rea} is higher.

\input{table/number}

\paragraph{Effect of expert number}
We vary the number of experts per routed layer
$N\in\{4,5,6,7\}$ while keeping top-2 expert selection to study how the number of experts affects performance. Table \ref{tab:number} shows that the overall performance improves from $N=4$ to $N=5$
and then decreases as $N$ increases to $6$ and $7$. $N=5$ achieves the best average results. A likely reason is that with too few experts, different tasks tend to rely on overlapping expert subsets, which is less effective for the rationale task that benefits from broader expert mixing; with too many experts, enlarging the expert pool does not yield further gains and can instead reduce the learning signal received by each expert. We therefore use $N=5$ in the main experiments.

%% file: table/ABSA.tex
\begin{table*}[!ht]
\caption{Benchmark evaluation of the proposed method against representative baselines on Rest14 and Lap14. The best score in each column is shown in bold. $^{\dag}$ indicates results are referred from \cite{laimulti}.}
   \label{tab:main-results}
    \centering    
%\scalebox{0.72}{
\begin{tabular}{llllllll}
\hline
\multicolumn{2}{c}{Model}                                                             & \multicolumn{3}{c}{Rest14}                                                                                   & \multicolumn{3}{c}{Lap14}                                                                                  \\
\cline{3-8}
\multicolumn{1}{c}{}                       & \multicolumn{1}{c}{}                     & \multicolumn{1}{c}{All$_A$}           & \multicolumn{1}{c}{All$_F$}           & \multicolumn{1}{c}{ISA$_A$}           & \multicolumn{1}{c}{All$_A$}           & \multicolumn{1}{c}{All$_F$}           & \multicolumn{1}{c}{ISA$_A$}      \\
\hline
\textbf{Conventional ABSA Baselines}                & \multicolumn{1}{c}{}                     & \multicolumn{1}{c}{}               & \multicolumn{1}{c}{}               & \multicolumn{1}{c}{}               & \multicolumn{1}{c}{}               & \multicolumn{1}{c}{}              & \multicolumn{1}{c}{}               \\
\multicolumn{1}{c}{BERT+SPC $^{\dag}$ \cite{BERT}}               & \multicolumn{1}{c}{BERT-base (110M)}     & \multicolumn{1}{c}{83.57}          & \multicolumn{1}{c}{77.16}          & \multicolumn{1}{c}{65.54}          & \multicolumn{1}{c}{78.22}          & \multicolumn{1}{c}{73.45}         & \multicolumn{1}{c}{69.54}          \\
\multicolumn{1}{c}{BERT+ADA $^{\dag}$ \cite{BERT-ADA}}               & \multicolumn{1}{c}{BERT-base (110M)}     & \multicolumn{1}{c}{87.14}          & \multicolumn{1}{c}{80.05}          & \multicolumn{1}{c}{65.92}          & \multicolumn{1}{c}{78.96}          & \multicolumn{1}{c}{74.18}         & \multicolumn{1}{c}{70.11}          \\
\multicolumn{1}{c}{RGAT $^{\dag}$ \cite{RGAT}}                   & \multicolumn{1}{c}{BERT-base (110M)}     & \multicolumn{1}{c}{86.6}           & \multicolumn{1}{c}{81.35}          & \multicolumn{1}{c}{67.79}          & \multicolumn{1}{c}{78.21}          & \multicolumn{1}{c}{74.07}         & \multicolumn{1}{c}{72.99}          \\
\multicolumn{1}{c}{BERT$_{Asp}$+CEPT $^{\dag}$ \cite{SCAPT}}           & \multicolumn{1}{c}{BERT-base (110M)}     & \multicolumn{1}{c}{87.5}           & \multicolumn{1}{c}{82.07}          & \multicolumn{1}{c}{67.79}          & \multicolumn{1}{c}{81.66}          & \multicolumn{1}{c}{78.38}         & \multicolumn{1}{c}{75.86}          \\
\multicolumn{1}{c}{BERT$_{Asp}$+SCAPT $^{\dag}$ \cite{SCAPT}}           & \multicolumn{1}{c}{BERT-base (110M)}     & \multicolumn{1}{c}{89.11}           & \multicolumn{1}{c}{83.79}          & \multicolumn{1}{c}{72.28}          & \multicolumn{1}{c}{82.76}          & \multicolumn{1}{c}{79.15}         & \multicolumn{1}{c}{77.59}          \\
\multicolumn{1}{c}{C3DA $^{\dag}$ \cite{C3DA}}                   & \multicolumn{1}{c}{BERT-base (110M)}     & \multicolumn{1}{c}{86.93}          & \multicolumn{1}{c}{81.23}          & \multicolumn{1}{c}{65.54}          & \multicolumn{1}{c}{80.61}          & \multicolumn{1}{c}{77.11}         & \multicolumn{1}{c}{73.57}          \\
\multicolumn{1}{c}{ISAIV $^{\dag}$ \cite{ISAIV}}                  & \multicolumn{1}{c}{BERT-base (110M)}     & \multicolumn{1}{c}{87.05}          & \multicolumn{1}{c}{81.4}           & \multicolumn{1}{c}{-}              & \multicolumn{1}{c}{80.41}          & \multicolumn{1}{c}{77.25}         & \multicolumn{1}{c}{-}              \\
\multicolumn{1}{c}{ABSA-ESA $^{\dag}$ \cite{ABSA-ESA}}               & \multicolumn{1}{c}{T5-base (220M)}       & \multicolumn{1}{c}{88.29}          & \multicolumn{1}{c}{81.74}          & \multicolumn{1}{c}{70.78}          & \multicolumn{1}{c}{82.44}          & \multicolumn{1}{c}{79.34}         & \multicolumn{1}{c}{80}             \\
\hline
\textbf{Inference-only Baselines}                   & \multicolumn{1}{c}{}                     & \multicolumn{1}{c}{}               & \multicolumn{1}{c}{}               & \multicolumn{1}{c}{}               & \multicolumn{1}{c}{}               & \multicolumn{1}{c}{}              & \multicolumn{1}{c}{}               \\
\multicolumn{1}{c}{GPT-5.4-mini}           & \multicolumn{1}{c}{}                     & \multicolumn{1}{c}{88.84}          & \multicolumn{1}{c}{81.92}          & \multicolumn{1}{c}{71.76}          & \multicolumn{1}{c}{82.39}          & \multicolumn{1}{c}{78.28}         & \multicolumn{1}{c}{73.29}          \\
\multicolumn{1}{c}{DeepSeek-V3.2}          & \multicolumn{1}{c}{}                     & \multicolumn{1}{c}{87.14}          & \multicolumn{1}{c}{76.14}          & \multicolumn{1}{c}{65.54}          & \multicolumn{1}{c}{79.62}          & \multicolumn{1}{c}{72.78}         & \multicolumn{1}{c}{60.57}          \\
\multicolumn{1}{c}{Llama-3.3-70b-instruct} & \multicolumn{1}{c}{}                     & \multicolumn{1}{c}{86.34}          & \multicolumn{1}{c}{74.91}          & \multicolumn{1}{c}{63.3}           & \multicolumn{1}{c}{80.56}          & \multicolumn{1}{c}{73.45}         & \multicolumn{1}{c}{59.43}          \\
\hline
\textbf{Instruction-based Fine-tuning   }           & \multicolumn{1}{c}{}                     & \multicolumn{1}{c}{}               & \multicolumn{1}{c}{}               & \multicolumn{1}{c}{}               & \multicolumn{1}{c}{}               & \multicolumn{1}{c}{}              & \multicolumn{1}{c}{}               \\
\multicolumn{1}{c}{Flan-T5 \cite{flan-t5}}                & \multicolumn{1}{c}{Flan-T5-base (250M)}  & \multicolumn{1}{c}{86.43}          & \multicolumn{1}{c}{77.45}          & \multicolumn{1}{c}{63.3}           & \multicolumn{1}{c}{80.41}          & \multicolumn{1}{c}{75.2}          & \multicolumn{1}{c}{71.43}          \\
\multicolumn{1}{c}{InstructABSA \cite{Instructabsa}} & \multicolumn{1}{c}{Flan-T5-base (250M)} & \multicolumn{1}{c}{85.62} & \multicolumn{1}{c}{75.52} & \multicolumn{1}{c}{62.17} & \multicolumn{1}{c}{81.82} & \multicolumn{1}{c}{77.72} & \multicolumn{1}{c}{76.57} \\

\multicolumn{1}{c}{THOR-prompt \cite{thor}}            & \multicolumn{1}{c}{Flan-T5-base (250M)}  & \multicolumn{1}{c}{86.7}           & \multicolumn{1}{c}{79.49}          & \multicolumn{1}{c}{65.92}          & \multicolumn{1}{c}{81.19}          & \multicolumn{1}{c}{77.42}         & \multicolumn{1}{c}{74.28}          \\
\multicolumn{1}{c}{THOR \cite{thor}}                   & \multicolumn{1}{c}{Flan-T5-base (250M)}  & \multicolumn{1}{c}{87.05}          & \multicolumn{1}{c}{80.09}          & \multicolumn{1}{c}{66.67}          & \multicolumn{1}{c}{81.66}          & \multicolumn{1}{c}{77.51}         & \multicolumn{1}{c}{74.29}          \\
\multicolumn{1}{c}{MT-ISA $^{\dag}$ \cite{laimulti}}                 & \multicolumn{1}{c}{Flan-T5-base (250M)}  & \multicolumn{1}{c}{88.21}          & \multicolumn{1}{c}{82.45}          & \multicolumn{1}{c}{70.41}          & \multicolumn{1}{c}{82.91}          & \multicolumn{1}{c}{79.86}         & \multicolumn{1}{c}{80.57}          \\
\multicolumn{1}{c}{Ours}                   & \multicolumn{1}{c}{Flan-T5-base (250M)}  & \multicolumn{1}{c}{88.23}          & \multicolumn{1}{c}{80.36}          & \multicolumn{1}{c}{69.16}          & \multicolumn{1}{c}{81.09}          & \multicolumn{1}{c}{75.88}         & \multicolumn{1}{c}{72.43}          \\
\multicolumn{1}{c}{Flan-T5 \cite{flan-t5}}                & \multicolumn{1}{c}{Flan-T5-large (780M)} & \multicolumn{1}{c}{88.75}          & \multicolumn{1}{c}{81}             & \multicolumn{1}{c}{71.91}          & \multicolumn{1}{c}{84.01}          & \multicolumn{1}{c}{79.58}         & \multicolumn{1}{c}{78.29}          \\
\multicolumn{1}{c}{InstructABSA \cite{Instructabsa}} & \multicolumn{1}{c}{Flan-T5-large (780M)} & \multicolumn{1}{c}{87.86} & \multicolumn{1}{c}{80.08} & \multicolumn{1}{c}{68.54} & \multicolumn{1}{c}{83.39} & \multicolumn{1}{c}{79.27} & \multicolumn{1}{c}{78.86} \\

\multicolumn{1}{c}{THOR-prompt \cite{thor}}            & \multicolumn{1}{c}{Flan-T5-large (780M)} & \multicolumn{1}{c}{88.13}          & \multicolumn{1}{c}{82.03}          & \multicolumn{1}{c}{72.29}          & \multicolumn{1}{c}{83.54}          & \multicolumn{1}{c}{80.3}          & \multicolumn{1}{c}{80}             \\
\multicolumn{1}{c}{THOR \cite{thor}}                   & \multicolumn{1}{c}{Flan-T5-large (780M)} & \multicolumn{1}{c}{88.21}          & \multicolumn{1}{c}{80.85}          & \multicolumn{1}{c}{70.41}          & \multicolumn{1}{c}{83.7}           & \multicolumn{1}{c}{81.01}         & \multicolumn{1}{c}{80.57}          \\
\multicolumn{1}{c}{Ours}                   & \multicolumn{1}{c}{Flan-T5-large (780M)} & \multicolumn{1}{c}{\textbf{90.21}} & \multicolumn{1}{c}{\textbf{84.74}} & \multicolumn{1}{c}{\textbf{75.9}} & \multicolumn{1}{c}{\textbf{85}} & \multicolumn{1}{c}{\textbf{81.63}} & \multicolumn{1}{c}{\textbf{82.67}} \\

\hline
\end{tabular}
%}
\end{table*}

%% file: table/ablation.tex
\begin{table}[!ht]
\caption{Ablation results on Rest14 and Lap14, isolating the effects of multi-task learning and MoE architecture. The best score in each column is shown in bold.}
\label{tab:ablation}
\centering
%          \scalebox{0.75}{

\begin{tabular}{lllllll}
\hline
                                   & \multicolumn{3}{c}{Rest14}                                                                                   & \multicolumn{3}{c}{Lap14}                                                                                  \\
& {All$_A$}           & \multicolumn{1}{c}{All$_F$}           & \multicolumn{1}{c}{ISA$_A$}           & \multicolumn{1}{c}{All$_A$}           & \multicolumn{1}{c}{All$_F$}           & \multicolumn{1}{c}{ISA$_A$}                         \\
\hline
\multicolumn{1}{c}{Ours}           &  \multicolumn{1}{c}{\textbf{90.21}} & \multicolumn{1}{c}{\textbf{84.74}} & \multicolumn{1}{c}{\textbf{75.9}} & \multicolumn{1}{c}{\textbf{85}} & \multicolumn{1}{c}{\textbf{81.63}} & \multicolumn{1}{c}{\textbf{82.67}} \\
\multicolumn{1}{c}{Ours (w/o MTL)} & \multicolumn{1}{c}{88.57}          & \multicolumn{1}{c}{81.37}          & \multicolumn{1}{c}{70.41}          & \multicolumn{1}{c}{84.17}          & \multicolumn{1}{c}{81}         & \multicolumn{1}{c}{80.57}          \\
\multicolumn{1}{c}{Ours (w/o MoE)} & \multicolumn{1}{c}{88.84}          & \multicolumn{1}{c}{81.34}          & \multicolumn{1}{c}{71.16}          & \multicolumn{1}{c}{83.54}          & \multicolumn{1}{c}{80.26}         & \multicolumn{1}{c}{81.71}          \\

\hline
\end{tabular}
%}
\end{table}

%% file: table/number.tex
\begin{table}[!ht]
\caption{Effect of expert number under the 
same experimental settings and top-2 expert selection. Best results are highlighted in bold. $N$ refers to the number of experts per routed layer.}
\label{tab:number}
\centering
%          \scalebox{0.75}{

\begin{tabular}{llllllll}
\hline
\multicolumn{1}{c}{\multirow{2}{*}{$N$}} & \multicolumn{3}{c}{Rest}                                                                & \multicolumn{3}{c}{Lap}                                                                 & \multicolumn{1}{c}{\multirow{2}{*}{\begin{tabular}[c]{@{}c@{}}Overall Mean \\ of Six Metrics\end{tabular}}} \\
\cline{2-7}
\\
\multicolumn{1}{c}{}                               &\multicolumn{1}{c}{All$_A$}           & \multicolumn{1}{c}{All$_F$}           & \multicolumn{1}{c}{ISA$_A$}           & \multicolumn{1}{c}{All$_A$}           & \multicolumn{1}{c}{All$_F$}           & \multicolumn{1}{c}{ISA$_A$}       & \multicolumn{1}{c}{}                                             \\
\hline
\multicolumn{1}{c}{4} & \multicolumn{1}{c}{89.55}          & \multicolumn{1}{c}{83.38}          & \multicolumn{1}{c}{74.32}          & \multicolumn{1}{c}{84.80}          & \multicolumn{1}{c}{81.21}          & \multicolumn{1}{c}{82.29}          & \multicolumn{1}{c}{82.59}          \\
\multicolumn{1}{c}{5} & \multicolumn{1}{c}{\textbf{90.21}} & \multicolumn{1}{c}{\textbf{84.74}} & \multicolumn{1}{c}{\textbf{75.90}} & \multicolumn{1}{c}{85.00}          & \multicolumn{1}{c}{81.63}          & \multicolumn{1}{c}{\textbf{82.67}} & \multicolumn{1}{c}{\textbf{83.36}} \\
\multicolumn{1}{c}{6} & \multicolumn{1}{c}{89.59}          & \multicolumn{1}{c}{83.47}          & \multicolumn{1}{c}{74.2}           & \multicolumn{1}{c}{\textbf{85.06}} & \multicolumn{1}{c}{\textbf{81.85}} & \multicolumn{1}{c}{82.48}          & \multicolumn{1}{c}{82.77}          \\
\multicolumn{1}{c}{7} & \multicolumn{1}{c}{89.02}          & \multicolumn{1}{c}{82.11}          & \multicolumn{1}{c}{71.54}          & \multicolumn{1}{c}{84.48}          & \multicolumn{1}{c}{81.12}          & \multicolumn{1}{c}{81.14}          & \multicolumn{1}{c}{81.57}          \\

\hline
\end{tabular}
%}
\end{table}

%% file: 5_conclusion.tex
\section{Conclusion}
\label{sec:conclusion}

In implicit aspect-level sentiment, sentiment toward an aspect is often conveyed through events rather than explicit opinion words. Polarity supervision alone is not enough to teach a model how an event implies an attitude toward an aspect. We introduce cognitive appraisal reasoning and implicit sentiment detection as auxiliary tasks, so that polarity prediction is trained together with signals about why the sentiment holds and whether the evidence is explicit or implicit. We then address the sharing problem introduced by the multi-task formulation with task-routed mixture-of-experts layers. Instead of forcing all objectives to pass through the same feed-forward networks, the model learns task-conditioned expert mixtures. The task-separated routing objective further encourages the routing distributions of different tasks to remain distinguishable. Performance gains in the benchmarks, including improvements in the implicit subset, demonstrate the effectiveness of the proposed method.